\documentclass[10pt,twocolumn,letterpaper]{article}

\usepackage[pagenumbers]{wacv} %

\usepackage{colortbl}                     %
\usepackage{pifont}                       %

\usepackage{dsfont}                        %
\newcommand{\indicator}{\mathds{1}}
\usepackage{tikz}                          %
\usepackage{dblfloatfix}                  %

\definecolor{wacvblue}{rgb}{0.21,0.49,0.74}
\usepackage[pagebackref,breaklinks,colorlinks,allcolors=wacvblue]{hyperref}

\def\wacvPaperID{2329} %
\def\confName{WACV}
\def\confYear{2027}

\title{Match the Amplitude, Not the Layout: Phase-Aware Distillation\\from RGB Foundation Models to Infrared Detection}

\author{
Keval Thaker$^{1,2}$ \and
Venkatraman Narayanan$^{1,3}$ \and
Abdalmalek Abu-Raddaha$^{1}$ \and
Samir Rawashdeh$^{1}$\\
$^{1}$University of Michigan--Dearborn\\
$^{2}$Torc Robotics\\
$^{3}$Automated Driving, Qualcomm Technologies, Inc.\\
{\tt\small \{tkeval, venn, abdmalek, srawa\}@umich.edu}
}
\begin{document}
\maketitle

\begin{abstract}
Cross-modal distillation measures a single similarity per frequency band, observes that the high band is near zero, and concludes that it should be aligned loosely. That inference conflates the \emph{kind} of disagreement with its degree. We decompose the agreement between a frozen RGB foundation teacher and an infrared (IR) student into two components that a scalar similarity confounds: \emph{phase}, which encodes the spatial position of structure, and \emph{amplitude}, which encodes its magnitude. Measured on several hundred registered pairs prior to any training, the two components diverge: for periods finer than $60$\,px amplitude agreement remains at $0.52$ while phase coherence falls to $0.06$, indistinguishable from chance. What ceases to transfer at fine scales is spatial location, not magnitude. The high band therefore requires a different \emph{quantity} rather than a smaller weight: we supervise the low band as a complex-valued feature and the high band in amplitude alone. The cut-off is not tuned on accuracy but derived as the radius at which teacher content ceases to be linearly recoverable from the student, yielding $r_c{=}0.47$. A single distilled IR backbone, reused without re-distillation, attains $64.1$ mAP$_{50}$ on KAIST Person ($+2.4$ over the DINOv2 baseline) and initializes FLIR detection ($+2.1$ mAP$_{50}$) and MFNet segmentation ($+1.85$ mIoU); the same objective also improves a ResNet-50 student ($+1.0$ mAP$_{50}$). All reported results use a DINOv2 ViT-Large teacher. Code and the distilled backbone are available at
\href{https://github.com/tkeval/freq_decoupled_kd}{\mbox{\nolinkurl{github.com/tkeval/freq_decoupled_kd}}}.
\end{abstract}

\section{Introduction}
\label{sec:intro}

Infrared (IR) perception underpins night-time driving~\cite{hwang2015kaist}, low-light surveillance~\cite{jia2021llvip} and search-and-rescue, yet IR detectors lag RGB for representational, not architectural, reasons. DINOv2~\cite{oquab2024dinov2} and CLIP~\cite{radford2021clip} give RGB encoders broad semantic coverage; thermal has no equivalent, with labeled data one to two orders of magnitude smaller and different imaging physics (emitted long-wave rather than reflected short-wave radiation~\cite{tattersall2016infrared}).

Cross-modal knowledge distillation (KD) --- an RGB teacher supervising an IR student on registered pairs --- is the natural remedy, but standard recipes fail~\cite{sikdar2025ogpnet}. Distilling RGB DINOv2 into an IR student on KAIST, uniform full-band feature matching reaches $61.1$ mAP$_{50}$ and response-level distillation $58.8$, both \emph{below} the $61.7$ undistilled baseline; per-token cosine alignment recovers only $+0.4$. This is diagnostic: a signal that makes the student worse contains an actively harmful component, and the harm scales with how much of the map is forced. The question is not whether to distil but which part: the teacher encodes appearance cues (color, illumination, reflectance) with no thermal counterpart, mixed into the same tensor as the geometry that does.

\textbf{The received answer is to relax the high band, and it conflates the kind of disagreement with its degree.} Cross-modal frequency distillation reports a single similarity per band, finds the high band near zero~\cite{liu2026fdcmkd}, and downweights it accordingly. A lumped similarity, however, cannot identify \emph{what} disagrees. Two components may fail independently: structure may be displaced, which is expressed in \emph{phase}, or its magnitude may differ, which is expressed in \emph{amplitude}. \Cref{sec:diagnosis} measures each against a null on frozen encoders, and the two are found to diverge: in the high band amplitude agreement holds at $0.52$ while phase coherence falls to $0.06$, at chance. What ceases to transfer at fine scales is spatial location rather than magnitude; the near-zero lumped similarity reflects not an absence of shared content but a superposition of amplitude agreement and phase noise.

\textbf{FreqKD} therefore changes the \emph{quantity} supervised in each band, not the weight. At each matched block, features are channel-centered and token-normalized, transformed by a 2D fast Fourier transform (FFT), and split at the measured cut-off. The low band, where phase and amplitude both agree, is matched as a complex-valued feature; the high band, whose layout is at chance, is matched in amplitude alone, asking for comparable fine-scale energy in the same places without reproducing its pattern or sign. At patch $14$, normalized radius $0.50$ is a period of four tokens, or $56$ px, so the split separates region-scale layout from token-scale variation, not pixel texture, which the patch embedding has already integrated away. Stage~1 trains low-rank adaptation (LoRA) adapters~\cite{hu2022lora} on unlabeled pairs and merges them into one reusable checkpoint.

\noindent\textbf{Contributions.}
\textbf{(1)}~\emph{A reinterpretation of the high band.} Separating phase from amplitude against a measured null across five frozen-encoder depths (\Cref{sec:diagnosis}) reveals that fine-scale amplitude agreement survives where phase collapses, showing near-zero lumped similarity is not evidence of absent content.
\textbf{(2)}~\emph{A different objective, not only a smaller weight.} Our diagnosis prescribes which quantity to match per band, deriving the cut-off from linear recoverability rather than an accuracy sweep. No prompt, filter, mask, or shared classifier is trained~\cite{liu2026fdcmkd,zhang2024freekd}; remaining scalars have reported sensitivity (\Cref{sec:method_freqkd}).
\textbf{(3)}~\emph{A reusable thermal backbone.} A single Stage-1 checkpoint, without re-distillation, initializes KAIST and FLIR detection and MFNet segmentation, and transfers to a ResNet-50 student.
\textbf{(4)}~\emph{Verification of the mechanism.} Frequency-resolved CKA, band ablations, and size-stratified AP verify selective spectral alignment and shows the high band is harmful when supervising layout but beneficial when matching amplitude alone.

\section{Related Work}
\label{sec:related}

\noindent\textbf{Multispectral and thermal detection.}
Multispectral pedestrian detectors on KAIST~\cite{hwang2015kaist} fuse registered RGB and thermal streams via dual-branch CNNs or cross-modal transformers~\cite{li2018msdsrcnn,zhou2020mbnet,chen2022proben,fang2021cft}, requiring both sensors at deployment. In contrast, we target single-modality IR inference, leveraging paired RGB strictly as privileged supervision.

\noindent\textbf{Foundation models and IR transfer.}
Vision foundation models pretrained on RGB data---including DINOv2~\cite{oquab2024dinov2}, SAM~\cite{kirillov2023sam,ravi2024sam2}, and CLIP/SigLIP~\cite{radford2021clip,zhai2023sigmoid,tschannen2025siglip2}---lack thermal priors, and fine-tuning on small IR datasets degrades representations. Recent thermal approaches attach task-specific adapters to frozen RGB backbones or align multimodal embeddings~\cite{xu2025samamba,zhao2025shif,distillmatch2025,girdhar2023imagebind}. We instead study upstream cross-modal transfer, adapting a foundation encoder via LoRA~\cite{hu2022lora} in Stage~1 and merging it into a thermal backbone.

\noindent\textbf{Knowledge distillation across modalities.}
Classical logit~\cite{hinton2015distilling} and feature distillation~\cite{romero2015fitnets,yang2022fgd,yang2022mgd,ranzinger2024amradio,ranzinger2024phis} assume teacher features are wholly reproducible within a shared modality. Cross-modal extensions bridge RGB--text~\cite{huo2024c2kd}, RGB--D~\cite{ferrod2025crodino}, modality hallucination~\cite{medeiros2024hallucidet}, and optical-guided segmentation~\cite{sikdar2025ogpnet}. For RGB-to-IR transfer, prior works distil into tracking heads~\cite{zhang2023crossmodality,li2025mkd}, detector necks~\cite{kim2025contrastguided}, or decoupled spatial--spectral branches~\cite{zhang2025ssdc}, yet all apply uniform supervision across spatial frequencies.

\noindent\textbf{Frequency-domain distillation.}
Spectral methods either incorporate Fourier/wavelet layers into network architectures~\cite{qin2021fcanet,rao2021gfnet,park2022how} or partition features by frequency band during distillation: FreeKD~\cite{zhang2024freekd} learns semantic frequency prompts, Wavelet KD~\cite{zhang2022wavelet} emphasizes high-frequency wavelet subbands, DS$^2$D$^2$~\cite{gao2025ds2d2} decouples wavelet directional bands, and Frequency Attention~\cite{pham2024freqattn} learns global spectral filters. However, these methods rely on heuristic or learned band splits rather than measuring cross-modal spectral divergence prior to training.

\noindent\textbf{Relation to FD-CMKD.}
Most closely related, FD-CMKD~\cite{liu2026fdcmkd} computes a scalar cosine similarity per frequency band and down-weights the high band where similarity approaches zero. We challenge this inference: a lumped scalar confounds structural alignment (phase) with magnitude agreement (amplitude). Disentangling the two (\Cref{sec:diagnosis}) reveals that strong high-band amplitude correspondence is masked by phase noise, motivating an amplitude-only objective rather than a reduced weight, with a cut-off radius derived directly from linear recoverability.

\begin{figure*}[!t]
  \centering
  \includegraphics[width=\linewidth]{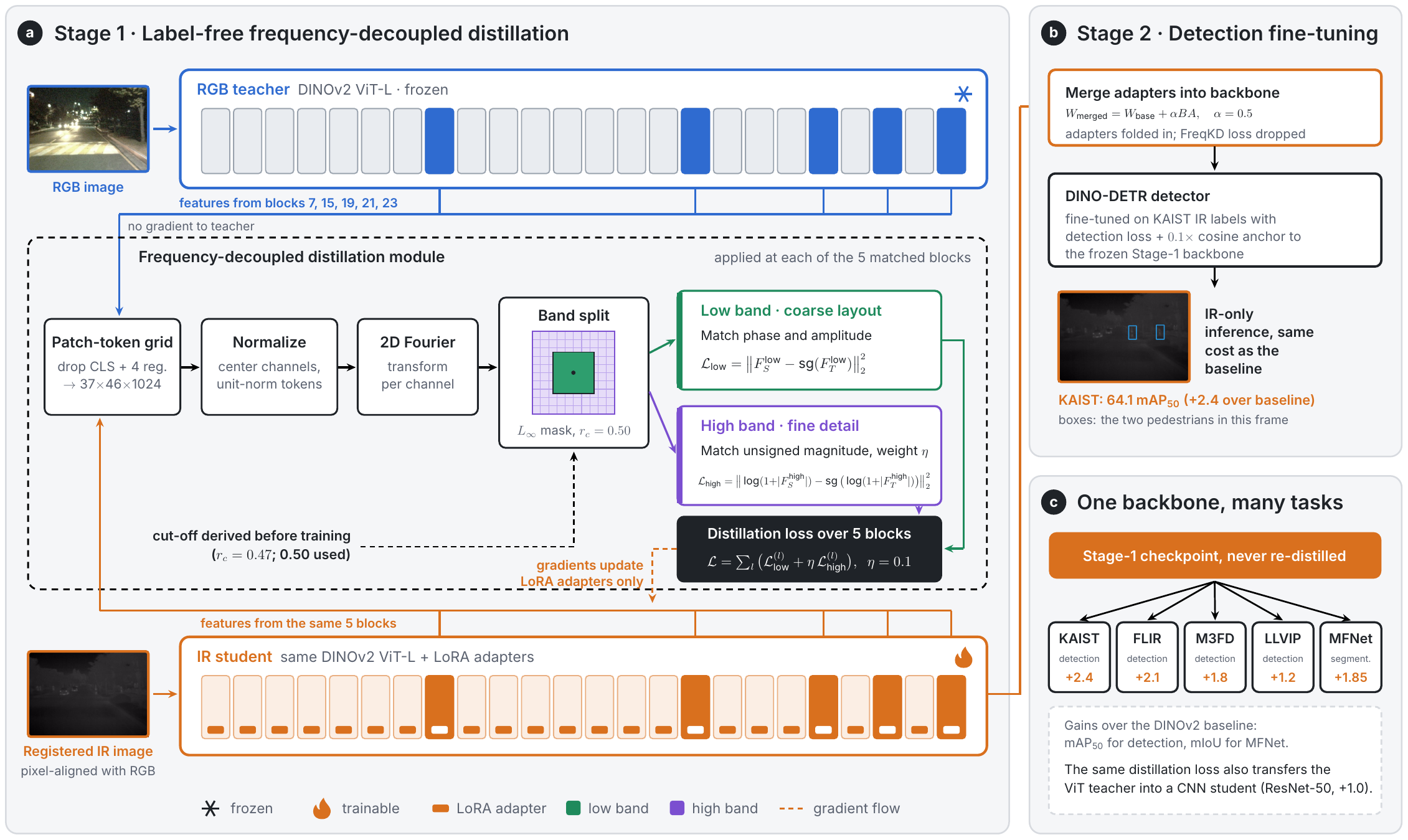}
  \caption{%
    \textbf{Overview of FreqKD.} \textbf{(a)}~Stage~1: a frozen RGB teacher and an IR student (the same DINOv2 ViT-L with rank-$64$ LoRA adapters) process registered pairs; only the adapters train. At each of the five matched blocks, features are channel-centered and token-normalized, 2D-FFT'd and split by the $L_\infty$ mask of \Cref{eq:mask} at $r_c{=}0.50$: the low band is supervised as a complex quantity (\Cref{eq:low}) and the high band by unsigned log-magnitude (\Cref{eq:high}), combined with $\eta{=}0.1$. \textbf{(b)}~Stage~2: the adapters merge with $\alpha{=}0.5$, and the backbone initializes a DINO-DETR detector fine-tuned with the detection loss and a light anchor to the frozen Stage-1 backbone. \textbf{(c)}~The same Stage-1 checkpoint initializes every downstream task without re-distillation.
  }
  \label{fig:architecture}
\end{figure*}

\section{Diagnosis: Where the RGB--IR Gap Lives}
\label{sec:diagnosis}

A band-asymmetric objective is justified only if three conditions hold, none of which a divergence establishes: the disagreement must persist once the energy carried by each band is accounted for; at fine scales it must concern the spatial \emph{position} of structure rather than its magnitude; and it must be \emph{irreducible} rather than merely unextracted. We evaluate all three on frozen encoders, prior to specifying any loss.

\noindent\textbf{Setup.}
We use $N{=}500$ registered KAIST pairs and both encoders \emph{before any training}: teacher and student are the \emph{same} public DINOv2 ViT-L checkpoint differing only in input --- the teacher takes RGB, the student the thermal image replicated to three channels under identical normalization --- so that the measurement isolates input modality on a \emph{fixed function} rather than comparing two trained networks. Inputs are $518{\times}644$\,px, giving $37{\times}46$ tokens at patch $14$. At each matched block $l\in\mathcal{L}{=}\{7,15,19,21,23\}$ we apply the loss's own preprocessing (\Cref{eq:standardize,eq:fft}), so that measurement and objective act on the same quantities up to the normalization axis, whose effect is audited below. Bins are $L_\infty$ rings matching the loss's mask (\Cref{eq:mask}), and all quantities are computed from \emph{complex} coefficients, so phase is retained throughout. Sec.~A of the supplementary derives each quantity and its null; Sec.~B gives the per-block table these means summarize, together with the controls below.

\noindent\textbf{Energy calibration.}
Comparing raw divergences across bands is the natural procedure and is nonetheless misleading. For band $b$ with mask $\mathbf{M}_b$, let $\mathcal{D}_b = \mathbb{E}\|\mathbf{M}_b\odot(\hat{\mathbf{F}}_T-\hat{\mathbf{F}}_S)\|_2^2$ denote the divergence and $\mathcal{E}_b = \sum_{X\in\{T,S\}}\mathbb{E}\|\mathbf{M}_b\odot\hat{\mathbf{F}}_X\|_2^2$ the total energy the two encoders place in that band. Expanding the square yields the \emph{energy-normalized divergence}
\begin{equation}
  \tilde{\mathcal{D}}_b = \frac{\mathcal{D}_b}{\mathcal{E}_b}
  = 1 - \frac{2\,\Re\langle\hat{\mathbf{F}}_{T,b},\hat{\mathbf{F}}_{S,b}\rangle}{\mathcal{E}_b} ,
  \label{eq:normdiv}
\end{equation}
with the inner product averaged over channels and samples. Being a ratio of energies, $\tilde{\mathcal{D}}_b$ is invariant to feature scale and to the normalization convention of the transform, both of which cancel, and it requires no empirical baseline: the cross term vanishes when the two bands share no structure, so $\tilde{\mathcal{D}}_b$ takes the value $0$ for identical features, $1$ under zero correspondence, and $2$ for anti-correlated ones. The same vanishing yields $\mathcal{D}_b{=}\mathcal{E}_b$ identically, so an uncalibrated band's divergence is simply its energy. A raw high-to-low ratio therefore cannot distinguish ``the modalities disagree here'' from ``this band carries more signal'', and since ViT features are strongly low-pass~\cite{park2022how}, the confound is substantial. We report $\tilde{\mathcal{D}}$ and the energy fractions separately (the low band holds $68\%$ of per-channel spectral energy at block~7, $79\%$ at block~23), never an uncalibrated ratio. Calibration does not merely preserve the asymmetry but produces it: the low band carries the majority of the energy yet disagrees least once normalized, whereas the two \emph{raw} divergences are near-equal.

\noindent\textbf{Phase and amplitude.}
The claim that low frequencies carry shared shape and position is a claim about spatial position, and spatial position is encoded in phase: translating a feature map alters its phase while leaving its magnitude unchanged, so no magnitude-only measurement can adjudicate it. With $a_X{=}|\hat{\mathbf{F}}_X(u,v)|$, amplitude agreement $A_b$ is the Pearson correlation of $a_T$ and $a_S$ over band $b$; phase coherence $\phi_b = \sum_b a_T a_S \cos\Delta\theta / \sum_b a_T a_S$, with $\Delta\theta = \angle\hat{\mathbf{F}}_T - \angle\hat{\mathbf{F}}_S$, weights each phase difference by the energy at that coefficient and ranges from $1$ for identical layout to $0$ at chance. So formulated, our account yields three simultaneous predictions, each of which can fail: $\phi_{\text{low}}\!\gg\!0$, $A_{\text{high}}\!\gg\!0$, $\phi_{\text{high}}\!\approx\!0$ (\Cref{fig:diagnosis}). All three hold at every matched block: $\phi_{\text{low}}{=}0.71$ and $A_{\text{high}}{=}0.52$ lie far above chance, while $\phi_{\text{high}}{=}0.06$ is at it --- precisely what a lumped divergence conceals. $\tilde{\mathcal{D}}_{\text{low}}{=}0.43$ lies well below the null of $1$, so the low band genuinely corresponds; $\tilde{\mathcal{D}}_{\text{high}}{=}0.93$ lies just below it, reproducing the prior observation that high-band agreement is near chance. That scalar, however, averages amplitudes that agree strongly against phases that do not, and returns approximately the null in either case. The near-zero lumped value is a \emph{superposition} of amplitude agreement and phase noise rather than an absence of shared fine-scale content, and its two components call for opposite treatment. The prescription follows: match complex coefficients where phase agrees and amplitude alone where it does not, since supervising layout at chance amounts to supervising noise.

\noindent\textbf{Recoverability and cut-off.}
Disagreement admits two causes that call for opposite responses: the content may be absent from the IR image, in which case relaxing the band is appropriate, or present but not yet captured, in which case supervision should be strengthened rather than relaxed. A divergence cannot distinguish them, so ``more divergent, therefore supervise less'' constitutes an argument only in conjunction with a second measurement. For band $b$ we fit a linear map $\mathbf{W}$ from student to teacher band-limited features on paired tokens and report held-out $R^2_b = 1 - \min_{\mathbf{W}}\mathbb{E}\|\mathbf{F}_{T,b}-\mathbf{W}\mathbf{F}_{S,b}\|_2^2 / \mathbb{E}\|\mathbf{F}_{T,b}-\bar{\mathbf{F}}_{T,b}\|_2^2$ against the floor obtained by refitting with the pairing shuffled; what is predictable is worth supervising even where it disagrees. Ring by ring, this yields the cut-off criterion
\begin{equation}
  r_c = \min\big\{\, r \;:\; R^2_{\text{pair}}(r) \le R^2_{\text{shuf}}(r) + \delta \,\big\} ,
  \label{eq:criterion}
\end{equation}
the smallest radius at which paired recoverability falls to the shuffled null within tolerance $\delta{=}0.02$; beyond it the teacher's content is not recoverable from IR. The criterion is fixed \emph{before} any detection metric is inspected, and yields $r_c{=}0.47$ ($60$\,px) on average, per-block $0.44$--$0.52$. Every block lies within one ring of that mean, which licenses a single shared cut-off, and their spread bounds the cost of sharing it at $0.3$ mAP$_{50}$. The accuracy sweep of \Cref{sec:method} peaks at $0.50$, one grid step from the derived value; trained models use $0.50$, the swept value nearest the derived one.

\noindent\textbf{Band scale.}
These radii admit an interpretation in terms of object size. A radius $\rho$ corresponds to a period of $2/\rho$ tokens, so at patch $14$, $\rho{=}0.50$ is $4$ tokens, or $56$\,px, and $\rho{=}0.75$ is $37$\,px. A KAIST \emph{reasonable} pedestrian is $\ge 55$\,px tall, so at $r_c{=}0.50$ the amplitude-only band begins at target-object scale. The derived cut-off therefore admits a direct reading: phase coherence collapses across this scale, reaching chance by about 47\,px, while amplitude agreement persists to 28\,px, so $r_c{=}0.47$ marks where the teacher's agreement about where structure sits breaks down while its agreement about how much of it there is continues. The split is region-scale \vs token-scale, not shape \vs micro-texture; nothing at this resolution supports claims about material texture.

\noindent\textbf{Controls.}
\emph{Shuffled pairs} give the chance floor for \Cref{eq:normdiv} and recoverability and verify the $\phi{=}0$ calibration; shuffled $\tilde{\mathcal{D}}$ is within $0.03$ of $1$ in every band. A \emph{grayscale-RGB teacher} --- luminance replicated to three channels through the student's pipeline --- separates imaging physics from the nuisance of a decolorized input; it recovers $38\%$ of the effect, so the larger component is attributable to imaging physics rather than to color removal. Two sensitivities audit \Cref{sec:method}'s choices --- $L_\infty$ \vs $L_2$ geometry at matched area ($r_c{=}0.50$ \vs $0.399$), per-channel \vs per-token normalization --- and neither moves $r_c$ by more than one ring.

\begin{figure}[tb]
  \centering
  \begin{tikzpicture}[x=7.15cm,y=1.95cm,line width=0.6pt]
    \draw[gray!45] (0,1) -- (1,1);
    \node[gray!60,font=\tiny,anchor=south east] at (1,1.01) {null};
    \draw[->] (0,0) -- (1.03,0) node[font=\scriptsize,anchor=north east,yshift=-1mm] {$\rho$};
    \draw[->] (0,0) -- (0,1.14);
    \foreach \y in {0,0.5,1} \draw (0,\y) -- (-0.012,\y) node[font=\tiny,anchor=east] {\y};
    \foreach \x/\pxl in {0.25/112,0.50/56,0.75/37,1.00/28}{
      \draw (\x,0) -- (\x,-0.03) node[font=\tiny,anchor=north] {\x};
      \node[font=\tiny,gray!70,anchor=south] at (\x,1.05) {\pxl};}
    \node[font=\tiny,gray!70,anchor=west] at (0.02,1.05) {period (px)};
    \draw[densely dotted,gray!80] (0.47,0) -- (0.47,1.10);
    \node[font=\tiny,anchor=south,inner sep=1pt] at (0.47,1.10) {$r_c$};
    \fill[gray!18] (0.05,0) rectangle (1,0.07);
    \node[font=\tiny,gray!70,anchor=west] at (0.06,0.035) {shuffled floor};
    \draw[red!75!black]  plot coordinates{(.05,.32)(.15,.36)(.25,.40)(.35,.45)(.45,.51)(.55,.70)(.65,.84)(.75,.89)(.85,.92)(.95,.94)};
    \draw[blue!70!black] plot coordinates{(.05,.80)(.15,.77)(.25,.72)(.35,.63)(.45,.42)(.55,.14)(.65,.08)(.75,.06)(.85,.05)(.95,.04)};
    \draw[green!45!black,densely dashed] plot coordinates{(.05,.93)(.15,.92)(.25,.90)(.35,.88)(.45,.85)(.55,.76)(.65,.67)(.75,.59)(.85,.53)(.95,.48)};
    \draw[orange!85!black,dash pattern=on 3pt off 1.5pt] plot coordinates{(.05,.86)(.15,.80)(.25,.68)(.35,.45)(.45,.13)(.55,.08)(.65,.07)(.75,.06)(.85,.06)(.95,.05)};
    \begin{scope}[shift={(0.60,0.48)},font=\tiny]
      \draw[red!75!black] (0,0.30) -- (0.055,0.30) node[anchor=west,inner sep=1.5pt] {$\tilde{\mathcal{D}}$};
      \draw[green!45!black,densely dashed] (0,0.20) -- (0.055,0.20) node[anchor=west,inner sep=1.5pt] {$A$ (amplitude)};
      \draw[blue!70!black] (0,0.10) -- (0.055,0.10) node[anchor=west,inner sep=1.5pt] {$\phi$ (phase)};
      \draw[orange!85!black,dash pattern=on 3pt off 1.5pt] (0,0) -- (0.055,0) node[anchor=west,inner sep=1.5pt] {$R^2_{\text{pair}}$};
    \end{scope}
  \end{tikzpicture}
  \caption{%
    Ring-resolved diagnosis, mean over the five matched blocks. Amplitude agreement $A$ stays far above the shuffled floor across the spectrum while phase coherence $\phi$ collapses into it near $\rho{\approx}0.5$: the high band's two components behave oppositely, which the single divergence $\tilde{\mathcal{D}}$, approaching its null, cannot show. $r_c$ is read off where $R^2_{\text{pair}}$ meets that floor (\Cref{eq:criterion}), not from any detection metric.}
  \label{fig:diagnosis}
\end{figure}

\section{Method}
\label{sec:method}

The teacher, student, input pipeline and matched blocks $\mathcal{L}{=}\{7,15,19,21,23\}$ are those of \Cref{sec:diagnosis}; sharing an architecture lets us distil block-for-block, student block $l$ supervised by teacher block $l$. Each block gives token features $\mathbf{F}_T^{(l)},\mathbf{F}_S^{(l)}$, reshaped (dropping class and register tokens) to maps in $\mathbb{R}^{H\times W\times d}$. \Cref{fig:architecture} summarizes the pipeline.

\subsection{Frequency-Decoupled Distillation Loss}
\label{sec:method_freqkd}

The loss is not an assumption about the modality gap but the objective prescribed by \Cref{sec:diagnosis}.

RGB and IR features differ systematically in mean and scale, so a direct comparison would be dominated by those differences rather than spatial pattern. We center each channel by its spatial mean and then $L_2$-normalize each token across channels: for $\mathbf{F}^{(l)}\in\{\mathbf{F}^{(l)}_T,\mathbf{F}^{(l)}_S\}$ and spatial position $(h,w)$,
\begin{equation}
  \tilde{\mathbf{F}}^{(l)}_{h,w} = \frac{\mathbf{F}^{(l)}_{h,w}-\boldsymbol{\mu}^{(l)}}{\max\!\big(\|\mathbf{F}^{(l)}_{h,w}-\boldsymbol{\mu}^{(l)}\|_2,\ \epsilon\big)} ,
  \label{eq:standardize}
\end{equation}
where $\mathbf{F}^{(l)}_{h,w}\in\mathbb{R}^{d}$ is the token at $(h,w)$, $\boldsymbol{\mu}^{(l)}\in\mathbb{R}^{d}$ holds the spatial mean of each channel, and $\epsilon{=}10^{-12}$. Normalizing each channel's map over space instead costs $0.2$ mAP$_{50}$, and \Cref{sec:diagnosis} reports the diagnosis under both.

For each block we apply a 2D FFT along the spatial dimensions and center the zero-frequency component,
\begin{equation}
  \hat{\mathbf{F}}^{(l)} = \mathrm{fftshift}\big(\mathrm{FFT}_{2D}(\tilde{\mathbf{F}}^{(l)})\big) \in \mathbb{C}^{H\times W\times d},
  \label{eq:fft}
\end{equation}
per channel. A binary mask $\mathbf{M}\in\{0,1\}^{H\times W}$ splits the centered spectrum by the $L_\infty$ norm of the frequency coordinate $(u,v)$, normalized by the per-axis Nyquist frequencies $u_{\max}{=}H/2$, $v_{\max}{=}W/2$:
\begin{equation}
  \begin{gathered}
  \rho(u,v) = \max\bigl(|u|/u_{\max},\; |v|/v_{\max}\bigr) \in [0,1] , \\[2pt]
  \mathbf{M}(u,v) = \indicator\bigl[\rho(u,v)\le r_c\bigr] ,
  \end{gathered}
  \label{eq:mask}
\end{equation}
so $\rho{=}0$ at the center and $\rho{=}1$ at every spectrum edge. The retained region is a centered rectangle covering about a fraction $r_c$ of each axis; at $r_c{=}0.50$ on the $37{\times}46$ token grid it keeps $18{\times}22$ coefficients ($23.3\%$). We use $L_\infty$ throughout, including the diagnostic bins of \Cref{sec:diagnosis}, so measurement and objective partition the spectrum identically. The geometries are \emph{not} interchangeable at equal $r_c$: with the $L_2$ radius normalized so that $\rho{=}1$ falls at the spectrum corner, a disc of radius $0.50$ retains $\pi/8{=}39.3\%$ of the coefficients in the continuous limit, and matching the $L_\infty$ area needs $r_c{=}0.399$. Cut-offs are therefore comparable only within a fixed geometry; \Cref{sec:diagnosis} reports the sensitivity at matched area. Masking gives $\hat{\mathbf{F}}_{\text{low}}=\mathbf{M}\odot\hat{\mathbf{F}}$ and $\hat{\mathbf{F}}_{\text{high}}=(1-\mathbf{M})\odot\hat{\mathbf{F}}$, whose inverse transforms are the band-limited features $\mathbf{F}_{\text{low}},\mathbf{F}_{\text{high}}$.

The cut-off is that derived by \Cref{eq:criterion}, $r_c{=}0.47$ ($60$\,px). Tuned on accuracy instead, mAP$_{50}$ over $r_c\in\{0.10,0.25,0.50,0.75\}$ peaks at $0.50$ (Sec.~D.2 of the supplementary), one grid step away. That the derived radius falls adjacent to the tuned optimum rather than upon it indicates that it constitutes a measurement rather than a restatement of the sweep. All trained models use $r_c{=}0.50$.

The bands are supervised differently because they transfer differently. On the low band both phase and amplitude transfer, so the student must reproduce the teacher's band-limited feature as a signed, complex quantity; squared error on the inverse-transformed band is, up to a constant, squared error on the complex low-band coefficients:
\begin{equation}
  \mathcal{L}_{\text{low}}^{(l)} = \big\|\mathbf{F}_{S,\text{low}}^{(l)} - \mathrm{sg}\big(\mathbf{F}_{T,\text{low}}^{(l)}\big)\big\|_2^2 ,
  \label{eq:low}
\end{equation}
with $\mathrm{sg}(\cdot)$ the stop-gradient for the frozen teacher. On the high band, phase is at chance (\Cref{fig:diagnosis}) while amplitude still agrees, so matching \emph{where} fine structure sits would supervise noise; we match its magnitude envelope under a log compression. Writing $\mathbf{A}_X^{(l)} = \log\big(1+\big|\mathbf{F}_{X,\text{high}}^{(l)}\big|\big)$, $X\in\{T,S\}$, with $|\cdot|$ and $\log$ elementwise,
\begin{equation}
  \mathcal{L}_{\text{high}}^{(l)} = \big\|\mathbf{A}_S^{(l)} - \mathrm{sg}\big(\mathbf{A}_T^{(l)}\big)\big\|_2^2 .
  \label{eq:high}
\end{equation}
\Cref{eq:high} is a \emph{texture-statistics} objective, not a texture-reproduction one: it asks the student to carry as much fine-scale energy as the teacher, in the same places to within sign, without placing the same pattern there. The compression tames the heavy-tailed amplitude distribution so a few high-energy tokens cannot dominate. Two consequences follow. First, the elementwise magnitude discards the \emph{sign} of the band-limited response, so a polarity-inverted edge counts as a match --- defensible in thermal imagery, where hot-on-cold and cold-on-hot contrast both occur, but a real loss of information. Second, discarding a spatial sign is the spatial-domain \emph{analog} of discarding spectral phase, not the same operation. A sign-preserving variant scores 63.8 and an exact spectral-amplitude objective 64.3: the sign costs little, and matching amplitude in the spectrum directly is marginally better.

The terms are combined across blocks with a fixed coefficient $\eta{=}0.1$ downweighting the high band,
\begin{equation}
  \mathcal{L}_{\text{FreqKD}} = \sum_{l\in\mathcal{L}} \big( \mathcal{L}_{\text{low}}^{(l)} + \eta\,\mathcal{L}_{\text{high}}^{(l)} \big) .
  \label{eq:freqkd}
\end{equation}

\noindent\textbf{What is derived and what is chosen.}
The band split is derived: geometry, cut-off, and the pairing of a complex with an amplitude-only objective all follow from \Cref{sec:diagnosis}. The remaining scalars are selected conventionally on validation data, and we report their sensitivity rather than claim a tuning-free method. The high-band weight is consequential ($\eta{=}1.0$ costs $2.1$ mAP$_{50}$; raw MSE in place of the compression costs $3.6$) and the LoRA merge scale more so: $\alpha$ spans $60.4$ to $64.1$ over $\{0.25,0.5,0.75,1.0\}$, a range wider than the $+2.4$ headline gain. FreqKD is therefore not parameter-free; what is new is that the split it applies is measured rather than posited.

\subsection{LoRA-Adapted Two-Stage Training}
\label{sec:method_twostage}

Stage~1 pre-trains the backbone under $\mathcal{L}_{\text{FreqKD}}$ alone for $12$ epochs, with no detection head or labels. Every dense student parameter is frozen and LoRA adapters~\cite{hu2022lora} of rank $r{=}64$ are injected into the query--key--value and MLP projections, so only $\mathbf{A}\in\mathbb{R}^{r\times d}$, $\mathbf{B}\in\mathbb{R}^{d\times r}$ update: the pretrained RGB prior is preserved while adapting to thermal input. Stage~2 merges the factors with a scale $\alpha$,
\begin{equation}
  \mathbf{W}_{\text{merged}} = \mathbf{W}_{\text{base}} + \alpha\cdot \mathbf{B}\mathbf{A} ,
  \label{eq:merge}
\end{equation}
using $\alpha{=}0.5$ throughout (Sec.~D.3 of the supplementary), and fine-tunes the merged backbone as the initialization of a five-scale DINO-DETR detector~\cite{zhang2023dino_detr} for $12$ epochs under standard classification, $\mathrm{L}_1$ and GIoU losses, with the backbone learning rate reduced tenfold.

Stage~2 adds one light term to the standard detection objective: at each matched block, $0.1\,(1-\cos)$ between the backbone's features and those of a frozen copy of the merged Stage-1 backbone on the same thermal input, which keeps fine-tuning close to the Stage-1 representation; it uses no RGB input and is discarded at inference. Because the single Stage-1 checkpoint is reused unchanged downstream (\Cref{sec:exp_transfer}), gains reflect the pretrained representation rather than task-specific tuning. This also separates FreqKD from objectives applying one pressure across the whole spectrum: uniform feature matching asks the student to reproduce fine-scale structure whose layout the diagnosis places at chance, and response-level KD aligns predictions conditioned on RGB appearance cues with no IR counterpart.

\section{Experiments}
\label{sec:experiments}

We evaluate FreqKD on KAIST, on band and cut-off ablations, and on transfer across datasets, tasks and architectures; \Cref{sec:limitations} collects limitations and missing controls.

\subsection{Setup}
\label{sec:exp_setup}

\noindent\textbf{Models, data and metrics.} The student backbone is DINOv2 ViT-L with $4$ register tokens (\texttt{vit\_large\_patch14\_reg4\_dinov2.lvd142m}), which also serves as the frozen RGB teacher; the head is the five-scale DINO-DETR~\cite{zhang2023dino_detr}. Both stages use AdamW~\cite{loshchilov2019adamw}, $\mathrm{lr}{=}5{\times}10^{-5}$, weight decay $10^{-4}$, batch $2$ per GPU, $12$ epochs, a $0.1\times$ backbone lr multiplier and gradient clipping at L$_2$ norm $0.1$; Stage~1 trains rank-$64$ LoRA factors, Stage~2 fine-tunes the merged backbone with the head. KAIST~\cite{hwang2015kaist} gives $9{,}466$/$967$ train/val images (\texttt{instancesonly\_filtered\_all-02}), \emph{Person} only; FLIR ADAS~\cite{flir_adas} and MFNet~\cite{ha2017mfnet} ($784$/$392$/$393$ images at $480{\times}640$) are the transfer targets. We report COCO mAP$_{50}$, mAP, AP$_S$/AP$_M$/AP$_L$, the KAIST log-average miss rate MR$^{-2}$ (lower better) and mIoU, as percentages from single runs; the two headline rows are repeated over $3$ seeds, giving $\pm 0.3$ mAP$_{50}$ for the baseline ($61.7$) and $\pm 0.4$ for FreqKD ($64.1$), so the $+2.4$ gap is several times the observed seed spread.

\noindent\textbf{Protocol disclosure.} Published KAIST miss rates are \emph{not} interchangeable with ours. The literature evaluates the $2{,}252$-image sanitized \emph{test} set under the \emph{reasonable}/\emph{all} subsets with official ignore-region semantics; we evaluate our $967$-image \emph{val} split under COCO matching with no ignore regions, so images, annotation cleaning and matching rules all differ. Every row of \Cref{tab:kaist} therefore carries a protocol tag, and MR$^{-2}$ is computed under \emph{our} rules: with nothing ignored, every occluded and sub-threshold person counts against us, which places our IR students at 38--46. Re-scoring the same checkpoints with the official toolbox restores comparability: the baseline scores 13.4 and FreqKD 11.9 MR$^{-2}$ on the \emph{reasonable} subset, preserving the ordering on our split. Sec.~E of the supplementary gives the long form, including the distinction between the sanitized \emph{training} annotations~\cite{li2018msdsrcnn} and the improved \emph{test} annotations~\cite{liu2018kaist_testanno} that the single word ``sanitized'' usually conflates.

\subsection{KAIST Pedestrian Detection}
\label{sec:exp_kaist}

\begin{table}[t]
\centering
\caption{%
  KAIST \emph{Person} detection (\%), grouped by role, not as a single ranking. Tags: \texttt{test}~= official sanitized $2{,}252$-image test set, \texttt{val}~= our $967$-image split, \texttt{RGB+IR}~= both modalities at inference. ``---''~= not reported by that work. Gray: frozen RGB teacher on RGB input, an upper reference, not a competitor. FreqKD uses $r_c{=}0.50$, $\eta{=}0.1$; AP$_M$ is in the text.
}
\label{tab:kaist}
\small
\setlength{\tabcolsep}{3pt}
\resizebox{\linewidth}{!}{%
\begin{tabular}{llccccc}
\toprule
\textbf{Method} & \textbf{Protocol} & \textbf{mAP} & \textbf{mAP$_{50}$} & \textbf{MR$^{-2}$$\downarrow$} & \textbf{AP$_S$} & \textbf{AP$_L$} \\
\midrule
\multicolumn{7}{l}{\emph{(i) Prior work, as reported --- not re-run; protocols differ}} \\
\quad MSDS-RCNN~\cite{li2018msdsrcnn} & \texttt{test/RGB+IR} & --- & --- & 11.6 & --- & --- \\
\quad MBNet~\cite{zhou2020mbnet} & \texttt{test/RGB+IR} & --- & --- & 8.1 & --- & --- \\
\quad ProbEn~\cite{chen2022proben} & \texttt{test/RGB+IR} & --- & --- & 5.1 & --- & --- \\
\midrule
\multicolumn{7}{l}{\emph{(ii) Baselines under our protocol --- IR-only student, same recipe}} \\
\quad DINOv2, no KD~\cite{oquab2024dinov2} & \texttt{val/IR} & 22.8 & 61.7 & 42.1 & 16.1 & 37.6 \\
\quad MMD-KD~\cite{gretton2012mmd} & \texttt{val/IR} & 22.4 & 61.0 & 43.0 & 16.2 & 37.1 \\
\quad CORAL~\cite{sun2016coral} & \texttt{val/IR} & 21.9 & 62.2 & 41.6 & 14.2 & 36.1 \\
\quad DANN~\cite{ganin2016dann} & \texttt{val/IR} & 21.6 & 61.4 & 43.4 & 14.0 & 36.4 \\
\quad FGD~\cite{yang2022fgd} & \texttt{val/IR} & 22.3 & 62.4 & 41.7 & 15.8 & 37.0 \\
\quad MGD~\cite{yang2022mgd} & \texttt{val/IR} & 22.6 & 62.9 & 41.1 & 16.0 & 37.3 \\
\quad FreeKD~\cite{zhang2024freekd} & \texttt{val/IR} & 22.7 & 63.2 & 40.5 & 16.3 & 36.6 \\
\rowcolor{blue!8}
\quad \textbf{FreqKD (ours)} & \texttt{val/IR} & \textbf{22.9} & \textbf{64.1} & \textbf{38.4} & \textbf{16.6} & 35.7 \\
\midrule
\multicolumn{7}{l}{\emph{(iii) Ablations of our loss --- variants of FreqKD}} \\
\quad Uniform full-band MSE & \texttt{val/IR} & 22.1 & 61.1 & 42.8 & 15.4 & 37.2 \\
\quad Per-token cosine KD & \texttt{val/IR} & 22.5 & 62.1 & 41.9 & 15.9 & 37.4 \\
\quad Response-level KD & \texttt{val/IR} & 20.8 & 58.8 & 45.6 & 14.0 & 35.9 \\
\midrule
\rowcolor{gray!15}
\quad \emph{RGB teacher, RGB input}~\cite{oquab2024dinov2} & \texttt{val/RGB} & \textit{26.4} & \textit{68.0} & \textit{33.2} & \textit{19.8} & \textit{41.2} \\
\bottomrule
\end{tabular}%
}
\end{table}

Block~(i) quotes each method at its authors' protocol, which is deliberately not like-for-like: all three need registered RGB+IR at inference --- an easier problem than IR-only detection --- use the sanitized test set rather than our val split, and report miss rate only. Until IR-only variants of these fusion detectors are run under our single-modality setting we make no claim about their standing relative to FreqKD.
Block~(ii) re-trains competing objectives under our protocol and constitutes the principal comparison. Against the baseline --- the public DINOv2 ViT-L checkpoint fine-tuned on IR, same RGB pre-training, no distillation, $61.7$ mAP$_{50}$ --- two distribution-matching objectives at the same five blocks bracket it: MMD-KD $61.0$ ($-0.7$) and CORAL $62.2$ ($+0.5$), against $64.1$ ($+2.4$) for FreqKD. Global statistic matching is therefore neither reliably helpful nor sufficient: a single distributional pressure cannot express a split that only frequency-resolved statistics reveal (\Cref{sec:diagnosis}). The closest competitor is the other spectral method, FreeKD at 63.2, so our margin over the strongest spectral baseline is 0.9 mAP$_{50}$; the modest size of that margin is itself informative, since what separates the two is the band-wise choice of \emph{quantity} rather than frequency decomposition per se. Block~(iii) ablates our own objective: uniform full-band MSE $61.1$ ($-0.6$), per-token cosine $62.1$ ($+0.4$), response-level KD $58.8$ ($-2.9$) --- spectrally blind objectives do not help and response-level transfer hurts.

\noindent\textbf{What the teacher contributes.} Two controls decompose the gain. Running Stage~1 with the RGB--IR pairing \emph{shuffled}, each thermal image distilled against a non-corresponding RGB frame, reaches 62.3; replacing the teacher altogether with a self-supervised objective at matched budget --- the same $12$ LoRA epochs on the same thermal images --- reaches 62.1. Against $61.7$ for no distillation and $64.1$ for FreqKD, the $+2.4$ therefore splits into $+0.4$ from in-domain thermal adaptation, a further $+0.2$ from an unpaired foundation-model prior, and $+1.8$ from cross-modal correspondence itself. Three quarters of the gain therefore disappears once the teacher no longer describes the same scene, which distinguishes cross-modal transfer from generic regularization.

\noindent\textbf{Size and IoU strictness.} FreqKD gains $+0.5$ AP$_S$ ($16.1{\to}16.6$) and $+0.2$ AP$_M$ ($23.2{\to}23.4$; MMD-KD $22.7$, CORAL $22.4$) but loses $1.9$ AP$_L$ ($37.6{\to}35.7$). AP$_L$ falls under \emph{every} cross-modal objective we ran --- MMD-KD $-0.5$, CORAL $-1.5$, FreqKD $-1.9$ --- so the regression is not specific to our loss, though it is largest under it, and we offer no mechanism for it (\Cref{sec:limitations}). Averaged over IoU thresholds all four systems lie within $1.0$ mAP ($22.8$/$22.4$/$21.9$/$22.9$), so the $+2.4$ mAP$_{50}$ is a gain in loose-IoU detection rather than in box quality; mAP$_{75}$ is low for all of them --- 9.1/8.8/$8.4$/9.4 in the same order, against mAP$_{50}$ above $61$ --- so box quality on this split is weak for every method, ours included, and FreqKD's advantage does not extend to it.

\noindent\textbf{Day/night and teacher choice.} Day/night mAP$_{50}$: 63.9/58.2 baseline \vs 65.4/62.0 FreqKD, so the gain concentrates at night ($+$3.8) rather than by day ($+$1.5). Night is the regime in which thermal input least resembles the teacher's RGB training distribution, consistent with Stage~1 importing RGB structural priors rather than merely re-registering thermal input into a form the head prefers; the shuffled-pair control above is the direct test of that interpretation. Varying only the frozen teacher (student, recipe, split and schedule fixed), every RGB foundation model we tried beats the $61.7$ baseline: DINOv2 $64.1$, DINOv3 64.3, CLIP ViT-L~\cite{radford2021clip} 63.4, SigLIP2 63.1, SAM ViT-L 62.6, supervised ImageNet-21k ViT-L 62.9. The self-supervised teachers lead and the capacity-matched supervised ViT-L trails them by 1.2--1.4, so teacher capacity alone does not account for the gain: the claim widens from ``a DINOv2 teacher'' to ``RGB foundation-model teachers'', though the diagnosis of \Cref{sec:diagnosis}, on DINOv2 alone, stays narrower.

\subsection{Frequency Band and Cut-off Ablation}
\label{sec:exp_band_ablation}

Isolating each band: low-band-only MSE $62.6$ mAP$_{50}$ ($+0.9$ over the $61.7$ baseline), high-band-only strict MSE $58.4$ ($-3.3$), and the asymmetric combination (low MSE plus $0.1{\times}$ high logMSE) $64.1$ ($+2.4$) --- the high-band signal is harmful when its layout is supervised, useful when only its amplitude is. Strict MSE demands the teacher's high band pointwise and in polarity, whereas the log-amplitude form of \Cref{eq:high} asks only for comparable fine-scale energy in the same places, which is what survives the change of sensor. This is the strongest evidence we have that the \emph{split}, not merely the additional objective, produces the gain.

\noindent\textbf{Cut-off.} Sweeping $r_c\in\{0.10,0.25,0.50,0.75\}$ gives $62.7$/$63.1$/$64.1$/$62.3$ mAP$_{50}$: every value beats the baseline, with a maximum at $r_c{=}0.50$, so the result is not specific to one cut-off. Relaxing the shared cut-off is a sensitivity check, not a contribution --- learning band selection is what prior work already does~\cite{zhang2024freekd,pham2024freqattn} --- and changes little: per-layer radii from \Cref{sec:diagnosis} give 64.4 and a learnable $r_c$ 64.6, against $64.1$, so sharing one cut-off across layers costs roughly 0.3--0.5 mAP$_{50}$.

\noindent\textbf{Representation alignment.} Under FreqKD, teacher--student Centered Kernel Alignment (CKA)~\cite{kornblith2019cka} ($500$ KAIST val pairs, split at $r_c{=}0.50$; Sec.~D.4 of the supplementary) moves the bands in opposite directions: low-band CKA rises ($0.82{\to}0.91$), high-band falls ($0.13{\to}0.02$), whereas uniform feature KD raises high-band CKA to $0.15$ and does not improve detection. Full-feature CKA nevertheless halves ($0.65{\to}0.30$), which the two-band decomposition does not account for: the residual is band-energy reallocation, with FreqKD moving 7\% of per-channel spectral energy out of the high band into the low, which alters the channel covariance CKA is computed over without changing either band's internal alignment.

\noindent\textbf{Versus distribution matching.} A frequency-resolved diagnostic on cached features (Sec.~D.4 of the supplementary; fixed features, not a trained sweep) puts the teacher--student MMD~\cite{gretton2012mmd} slightly \emph{larger} in the low band than the high (ratio $0.74$): global matching does not separate the bands as FreqKD requires, consistent with MMD-KD failing to beat the baseline.

\subsection{Transfer Across Datasets, Tasks and Architectures}
\label{sec:exp_transfer}

\begin{figure}[t]
\centering
\includegraphics[width=\linewidth]{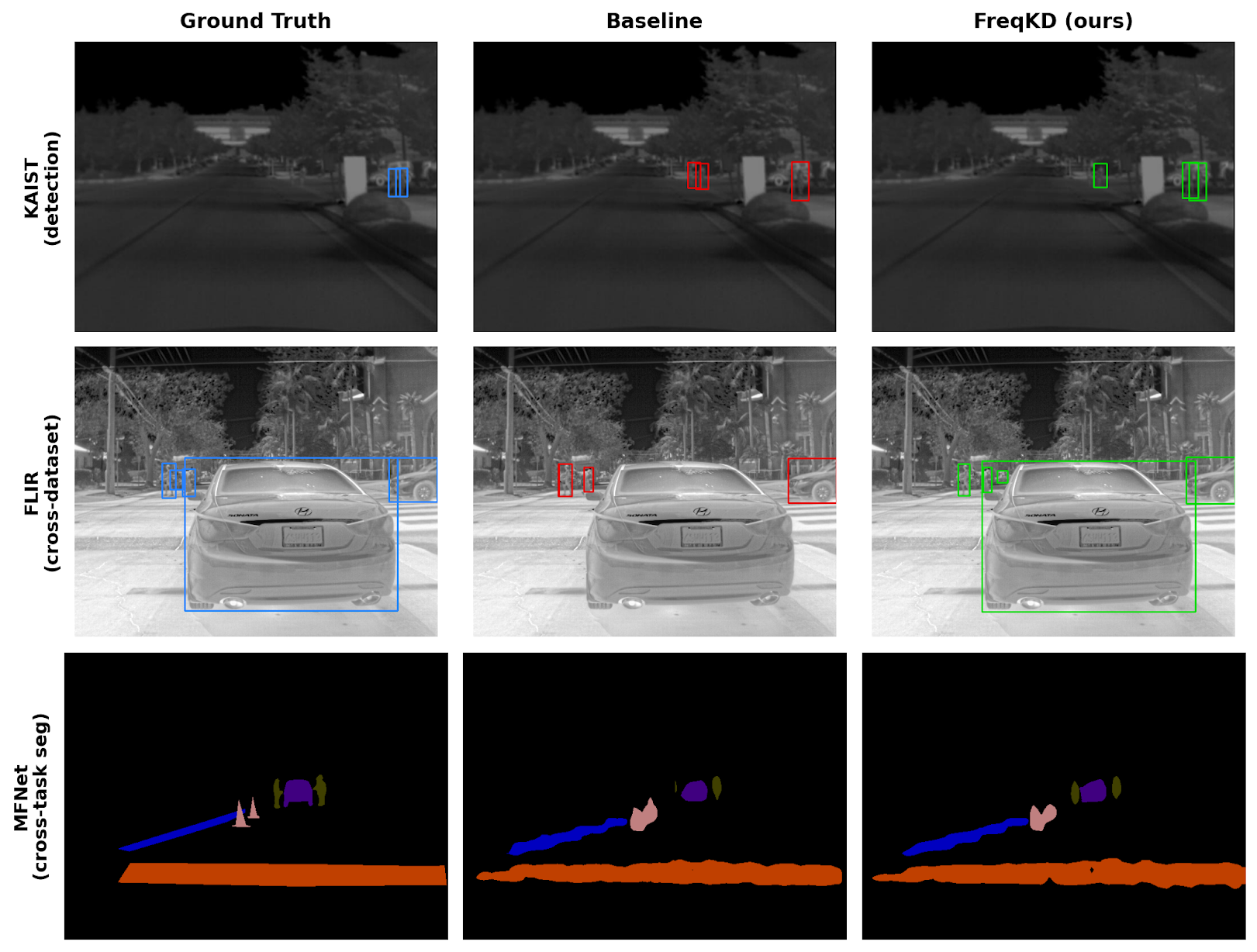}
\caption{%
  Qualitative results across the three transfer settings (GT, DINOv2 baseline, FreqKD). \emph{Top, KAIST:} the baseline misses pedestrians near the gate and fires on background; \emph{middle, FLIR:} FreqKD localizes the central vehicle and the partially visible side cars; \emph{bottom, MFNet:} cleaner \emph{person} and \emph{curve} masks.
}
\label{fig:qualitative}
\end{figure}

The single KAIST Stage-1 checkpoint is reused unchanged in all three settings (\Cref{fig:qualitative}; full tables in Sec.~D.5 of the supplementary). On FLIR ADAS it gains $+2.1$ mAP$_{50}$ ($70.5{\to}72.6$) but only $+0.2$ mAP ($34.9{\to}35.1$) over the public DINOv2 initialization --- the loose-IoU-only pattern of KAIST --- with gains on all three classes, the same small-object skew (AP$_S$ 12.6${\to}$14.1), and a wide margin over random init ($18.4$/$39.8$). Two further thermal sets move the same way with the same checkpoint: M3FD 71.2${\to}$73.0 ($+$1.8) and LLVIP~\cite{jia2021llvip} 88.4${\to}$89.6 ($+$1.2) mAP$_{50}$. Both arms start from the same DINOv2 checkpoint, so the gain is not RGB pre-training alone; it is, however, attributable to \emph{Stage~1 as a whole} rather than to cross-modal distillation specifically, since the arms differ in objective \emph{and} in $12$ epochs of LoRA adaptation on thermal images. The teacher-free control above separates them: matched-budget adaptation alone accounts for roughly a sixth of the KAIST gain, so the objective carries the rest here too.

MFNet tests \emph{task-agnosticism}, not segmentation performance: whether the Stage-1 representation stays useful once the detection head it was trained with is swapped for a dense predictor over a frozen backbone. Dataset and task both change, so the $+1.85$ mIoU ($46.95{\to}48.80$) depends on neither. One caveat: we average IoU over \emph{eight} foreground classes whereas the MFNet standard averages \emph{nine}, including the large and easily segmented background class; $48.80$ should therefore be compared only against our identically computed $46.95$ and not against published mIoU. The margin widens as supervision shrinks --- at $10$/$25$/$50\%$ of labels, 24.1/33.5/41.0 mIoU baseline \vs 28.4/36.9/43.2 FreqKD, i.e.\ $+$4.3/$+$3.4/$+$2.2 against $+1.85$ at full supervision --- so the Stage-1 backbone helps most where target labels are scarce, which is what makes it a reusable initialization rather than a detection-specific one.

The third setting tests the loss beyond transformers: a Faster R-CNN~\cite{ren2015fasterrcnn} ResNet-50 IR student, distilled from the DINOv2 ViT-L teacher into its four stages, gains $+1.0$ mAP$_{50}$ ($54.7{\to}55.7$), whereas a same-architecture ResNet-50 teacher leaves it unchanged --- the gain is not generic regularization from a distillation term. The two teachers differ in architecture, capacity and pre-training data simultaneously, so the claim is restricted to the weaker one that a large RGB foundation-model teacher benefits a CNN student where a same-architecture ImageNet teacher does not. A capacity-matched supervised ViT-L teacher separates these factors: it reaches 55.1, so capacity accounts for part of the gain and foundation pre-training for the remainder.

\subsection{Ablation Studies}
\label{sec:exp_ablation}

Remaining design choices are ablated on KAIST mAP$_{50}$ under the same $12{+}12$-epoch budget, so differences reflect design, not training length (Sec.~D.3 of the supplementary): high-band log compression matters (raw MSE $60.5$, below the $61.7$ baseline; undownweighted $\eta{=}1.0$ recovers only $+0.3$, i.e.\ $62.0$), the centered $L_2$ normalization of \Cref{eq:standardize} contributes $+1.4$ ($62.7$ \vs $64.1$), and all five blocks ($64.1$) beat every tested subset (best $63.2$). LoRA rank, merge scale ($\alpha{=}0.5$) and cost are there too; inference cost equals the baseline, as losses and adapters are discarded after merging.

\section{Limitations}
\label{sec:limitations}
\noindent\textbf{Large objects regress.} FreqKD loses $1.9$ $\text{AP}_L$ on KAIST (MMD-KD $-0.5$, CORAL $-1.5$), showing the drop is not specific to our loss, though largest under it and unaccounted for in this small bucket ($214$ boxes).

\noindent\textbf{Single measured teacher.} While gains hold across five other RGB teachers (\Cref{sec:exp_kaist}), the motivating phase/amplitude split is measured only on DINOv2 ViT-L.

\noindent\textbf{Fixed mask and protocol.} Using a fixed $L_\infty$ cut-off ($r_c=0.50$) across layers and datasets costs at most $0.5$ $\text{mAP}_{50}$ versus per-layer or learnable variants (\Cref{sec:exp_band_ablation}). Also, our KAIST split differs from the standard protocol (\Cref{sec:exp_setup}).

\noindent\textbf{Detection rate vs.\ box quality.} Gains concentrate at loose IoU on FLIR and KAIST: amplitude-only high-band matching discards fine layout cues needed for tight boxes.

\noindent\textbf{Residual gap.} Roughly $62\%$ of the RGB teacher gap ($68.0$ vs.\ $64.1$) remains, largely bounded by thermal imaging physics (\Cref{sec:diagnosis}).

\section{Conclusion}
\label{sec:conclusion}

Near-zero cross-modal similarity does not imply untransferable content. Separating phase and amplitude shows that fine-scale layout fails while magnitude agrees, requiring a shift in \emph{what} is matched. Our approach yields a thermal backbone reaching $64.1$ $\text{mAP}_{50}$ on KAIST ($+2.4$), which directly transfers without re-distillation to FLIR, MFNet, and a ResNet-50 student. This framework applies broadly whenever phase and amplitude are separately measurable.

{
    \small
    \bibliographystyle{ieeenat_fullname}
    \bibliography{main}
}

\end{document}